\documentclass[10pt,twocolumn,letterpaper]{article}

\usepackage[final]{cvpr}      %

\usepackage{graphicx}
\usepackage{amsmath}
\usepackage{amssymb}
\usepackage{booktabs}
\usepackage[usenames,dvipsnames]{xcolor}
\usepackage{url}
\usepackage{enumitem}
\usepackage{xspace}
\usepackage{paralist}
\usepackage{multirow}

\usepackage[pagebackref,breaklinks,colorlinks]{hyperref}

\makeatletter

\newcommand{\Rmnum}[1]{\expandafter\@slowromancap\romannumeral #1@}
\makeatother

\usepackage{bm}

\newif\ifshowcomments
\showcommentstrue %

\ifshowcomments
    \newcommand{\todo}[1]{\noindent\textcolor{BrickRed}{\textbf{[TODO:~}#1\textbf{]}}}
    \newcommand{\otmar}[1]{\noindent\textcolor{ForestGreen}{\textbf{[Otmar:~}#1\textbf{]}}}
    \newcommand{\xu}[1]{\noindent\textcolor{Salmon}{\textbf{[Xu:~}#1\textbf{]}}}
    \newcommand{\jie}[1]{\noindent\textcolor{Dandelion}{\textbf{[Jie:~}#1\textbf{]}}}
    \newcommand{\zijian}[1]{\noindent\textcolor{Purple}{\textbf{[Zijian:~}#1\textbf{]}}}
    \newcommand{\AS}[1]{\noindent\textcolor{BrickRed}{\textbf{[Adrian:~}#1\textbf{]}}}
    \newcommand{\MK}[1]{\noindent\textcolor{Dandelion}{\textbf{[Manuel:~}#1\textbf{]}}}
    \newcommand{\OH}[1]{{\color{blue}[OH: #1]}}
    \newcommand{\ag}[1]{{\color{cyan}[AG: #1]}}
    \newcommand{\pmnote}[1]{\PM{#1}}
    \newcommand{\oh}[1]{\OH{#1}}
    \newcommand{\zj}[1]{{\color{red}[zj: #1]}}
    \newcommand{\cg}[1]{{\color{magenta}[cg: #1]}}
    \newcommand{\JZ}[1]{{\color{red}[JZ: #1]}}
\else
    \newcommand{\todo}[1]{\unskip}
    \newcommand{\otmar}[1]{\unskip}
    \newcommand{\xu}[1]{\unskip}
    \newcommand{\jie}[1]{\unskip}
    \newcommand{\zijian}[1]{\unskip}
    \newcommand{\AS}[1]{\unskip}
    \newcommand{\MK}[1]{\unskip}
    \newcommand{\OH}[1]{\unskip}
    \newcommand{\ag}[1]{\unskip}
    \newcommand{\pmnote}[1]{\unskip}
    \newcommand{\oh}[1]{\unskip}
    \newcommand{\zj}[1]{\unskip}
    \newcommand{\cg}[1]{\unskip}
    \newcommand{\JZ}[1]{\unskip}

\fi    
    
\newcommand{\methodname}{PINA\xspace}
\newcommand{\suppmat}{Supp. Mat\xspace}

\newcommand{\figref}[1]{Fig.~\ref{#1}}
\newcommand{\tabref}[1]{Tab.~\ref{#1}}

\DeclareMathOperator*{\argmin}{arg\,min}

\definecolor{babyblue}{rgb}{0.54, 0.81, 0.94}

\usepackage[capitalize]{cleveref}
\crefname{section}{Sec.}{Secs.}
\Crefname{section}{Section}{Sections}
\Crefname{table}{Table}{Tables}
\crefname{table}{Tab.}{Tabs.}

\usepackage[accsupp]{axessibility}  

\begin{document}

\title{\methodname: Learning a Personalized Implicit Neural Avatar\\
from a Single RGB-D Video Sequence}

\author{Zijian Dong$^{*1}$ \quad Chen Guo$^{*1}$ \quad Jie Song$^{\dag1}$ \quad Xu Chen$^{1,2}$ \quad Andreas Geiger$^{2,3}$  \quad Otmar Hilliges$^{1}$ \\
 $^1$ETH Z{\"u}rich \quad 
 $^2$Max Planck Institute for Intelligent Systems, T{\"u}bingen \\
 $^3$University of T{\"u}bingen \\
}

\twocolumn[{%
\renewcommand\twocolumn[1][]{#1}%
\maketitle
\vspace{-4 em}
\begin{center}
    \captionsetup{type=figure}
   \includegraphics[width=\linewidth,trim=0 50 0 0]{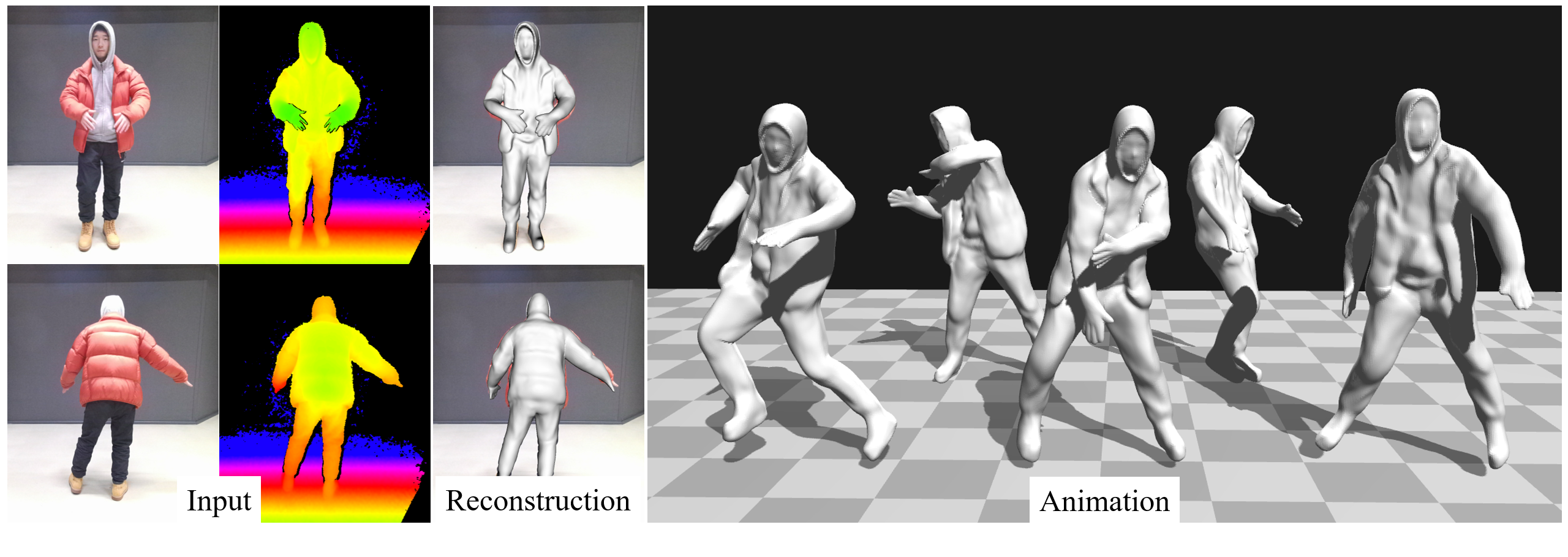}
    \captionof{figure}{We propose \methodname, a method to acquire personalized and animatable neural avatars from RGB-D videos. \emph{Left}: our method uses only a single sequence, captured via a commodity depth sensor. The depth frames are noisy and contain only partial views of the body. \emph{Middle}: Using global optimization, we fuse these partial observations into an implicit surface representation that captures geometric details, such as loose clothing. The shape is learned alongside a pose-independent skinning field, supervised only via depth observations. \emph{Right}: The learned avatar can be animated with realistic articulation-driven surface deformations and generalizes to novel unseen poses. 
    }\label{fig:teaser}
    
\end{center}%

}]

\def\thefootnote{*}\footnotetext{Equal contribution}
\def\thefootnote{\dag}\footnotetext{Corresponding author}

\newcommand{\figurePipeline}{

\begin{figure*}[t]
\includegraphics[width=\linewidth]{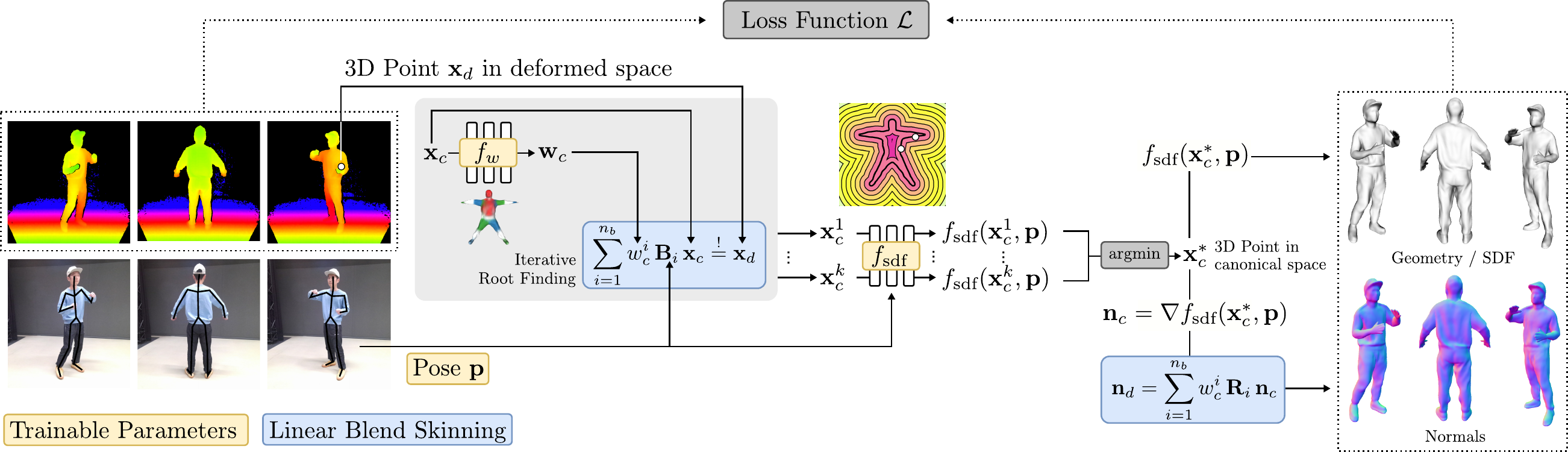}
\caption{\textbf{Method Overview.} Given input depth frames and human pose initializations inferred from RGB-D images, we first sample 3D points $\mathbf{x}_{d}$ on and off the human body surface in deformed (posed) space. Their corresponding canonical location $\mathbf{x}_{c}^*$ is calculated via iterative root finding~\cite{chen2021snarf} of the linear blend skinning constraint (here $\scriptstyle \overset{!}{=}$ denotes that we seek the top-$k$ roots $\mathbf{x}_c^{1:k}$, indicating the possible correspondences) and minimizing the SDF over these $k$ roots. Given the canonical location $\mathbf{x}_{c}^*$, we evaluate the SDF of $\mathbf{x}_{c}^*$ in canonical space, obtain its normal as the spatial gradient of the signed distance field and map it into deformed space using learned linear blend skinning. We minimize the loss $\mathcal{L}$ that compares these predictions with the input observations. Our loss regularizes off-surface points using proxy geometry and uses an Eikonal loss to learn a valid signed distance field $f_{\text{sdf}}$.}
\label{fig:pipeline}
\end{figure*}
}

\newcommand{\figureBUFF}{

\begin{figure*}[t]
\includegraphics[width=\linewidth,trim=0 10 0 2,clip]{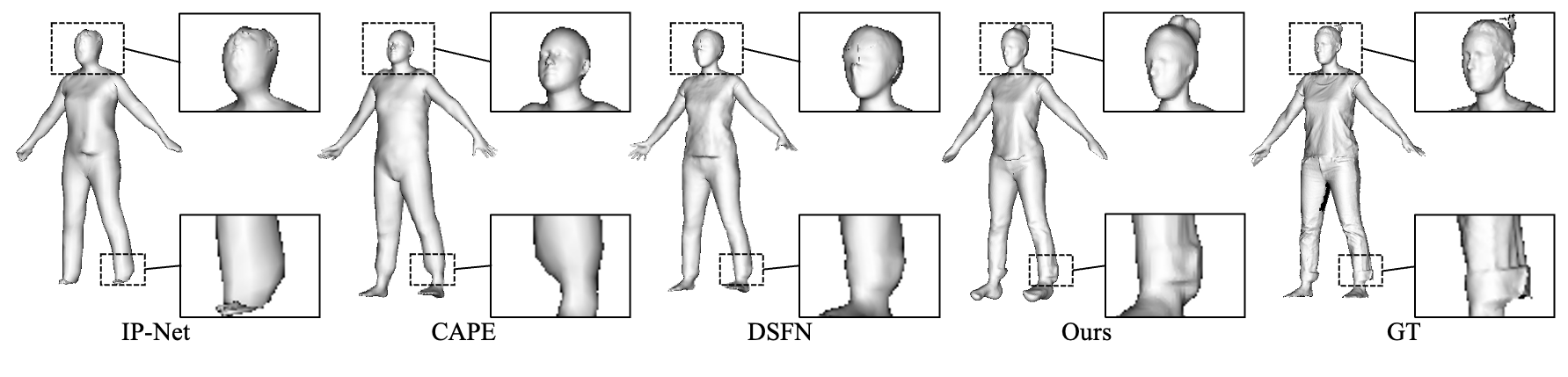}
\vspace{-2em}
\caption{\textbf{Qualitative reconstruction comparisons on BUFF.} Our method reconstructs better details and generates less artifacts compared to IP-Net. The implicit shape representation enables accurate reconstruction of complex geometry (hair, trouser heel) compared to methods with explicit representations, i.e., CAPE and DSFN.}

\label{fig:buff_result}
\end{figure*}
} 
\newcommand{\figureCape}{

\begin{figure*}[t]
\includegraphics[width=\linewidth,trim=0 15 0 0,clip]{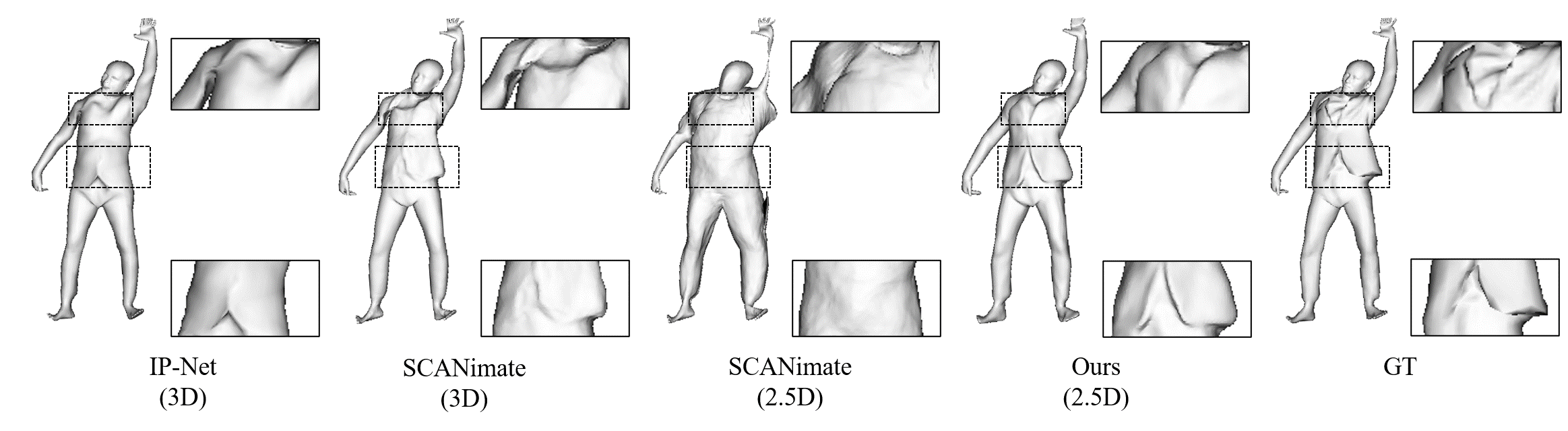}

\caption{\textbf{Qualitative animation comparison on CAPE.} IP-Net produces unrealistic animation results potentially due to overfitting and wrong skinning weights. The deformation field of SCANimate is defined in the deformed space and thus limits its generalization to unseen poses. This is made worse in SCANimate (2.5D) which only uses partial point clouds as input. In contrast, our method solves this problem naturally via joint optimization of skinning field and shape in canonical space.}

\label{fig:cape_result}
\end{figure*}
} 

\newcommand{\figureablationopt}{

\begin{figure}[t]
\raggedleft
\includegraphics[width=\linewidth,trim=0 3 0 0,clip]{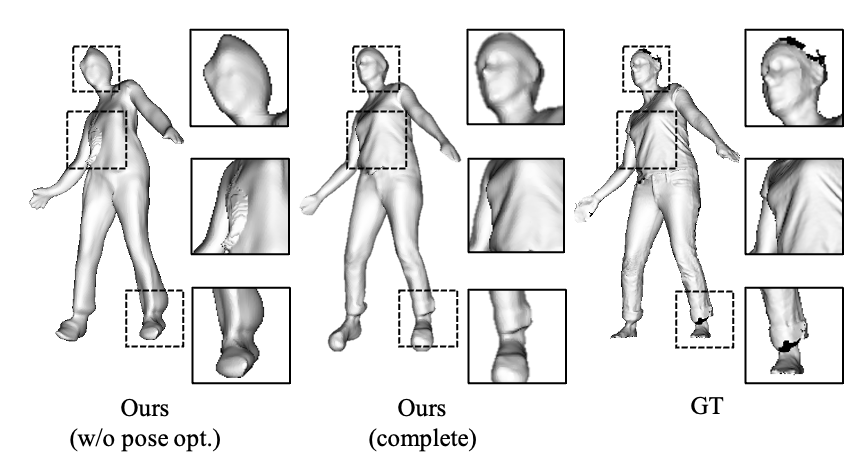}
\vspace{-2em}
\caption{\textbf{ Qualitative ablation (BUFF)}. Joint optimization corrects pose estimates and achieves better reconstruction quality.}

\label{fig:ablation_opt}
\end{figure}
} 

\newcommand{\figureablationdeform}{

\begin{figure}[t]
\raggedleft
\includegraphics[width=\linewidth,trim=0 1 0 10,clip]{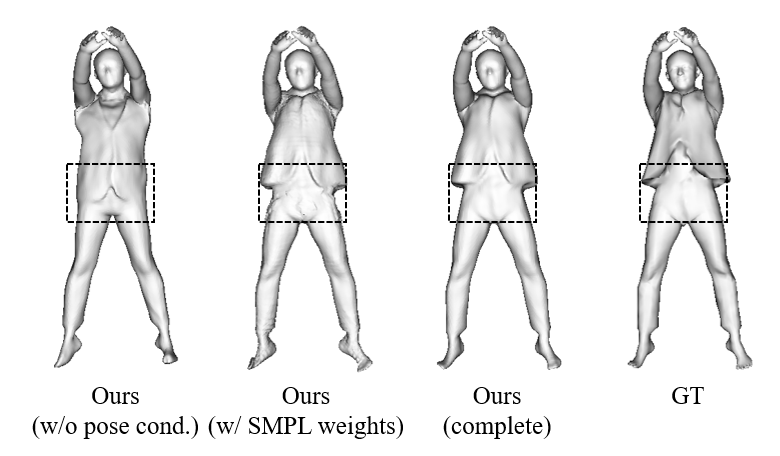}
\vspace{-2em}
\caption{\textbf{Qualitative ablation (CAPE).} Without conditioning our shape network on poses, the static network cannot represent pose-dependent surface details including deformations on necklines and bottom hems. Further, the lack of learned skinning weights results in noisy surfaces.}

\label{fig:ablation_deform}
\end{figure}
} 

\newcommand{\figurecompreal}{

\begin{figure}[t]
\begin{center}
\raggedleft
\includegraphics[width=\linewidth,trim=0 10 0 0,clip]{supplementary/figures/comp_real.png}
\end{center}
   \caption{\textbf{Qualitative evaluation of energy functions.} Dropping the off-surface loss $E_{\text{off}}$ results in many artifacts outside the human body. Ignoring the normal loss term $E_{\text{normal}}$ leads to less detailed reconstructions.}
\label{fig:comp_real}
\end{figure}
}

\newcommand{\figureablationloss}{

\begin{figure}[t]
\begin{center}
\raggedleft
\includegraphics[width=\linewidth,trim=0 10 0 0,clip]{figures/ablation_loss.png}
\end{center}
   \caption{\textbf{Qualitative evaluation of energy functions.} Dropping the off-surface loss $E_{\text{off}}$ results in many artifacts outside the human body. Ignoring the normal loss term $E_{\text{normal}}$ leads to less detailed reconstructions.}
\label{fig:ablation_loss}
\end{figure}
} 

\newcommand{\figureQualRes}{

\begin{figure*}
\includegraphics[width=\linewidth,trim=0 15 0 0,clip]{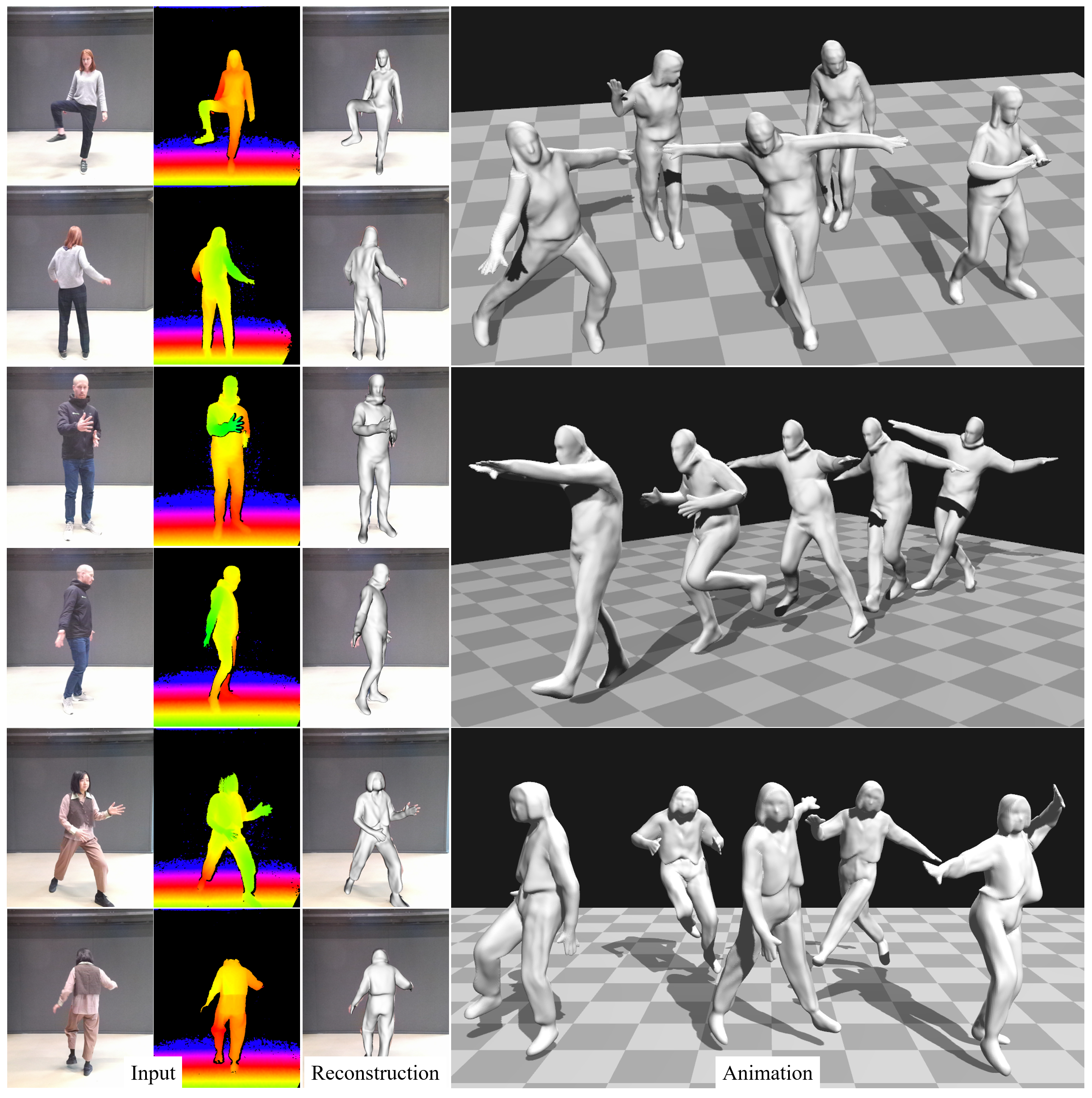}
\vspace{-1em}
\caption{\textbf{RGB-D results.} We show qualitative results of our method from real RGB-D videos. Each subject has been recorded for 2-3 min (left). From noisy depth sequences (cf. head region bottom row), we learn shape and skinning weights (reconstruction), by jointly fitting the parameters of the shape and skinning network and the poses. We use unseen poses from \cite{aist-dance-db,AMASS:ICCV:2019,CAPE:CVPR:20} to animate the learned character.}

\label{fig:qualitative_result}
\end{figure*}
}

\begin{abstract}
\vspace{-0.2em}
\vspace{-1em}
We present a novel method to learn \textbf{P}ersonalized \textbf{I}mplicit \textbf{N}eural \textbf{A}vatars (\methodname) from a short RGB-D sequence. 
This allows non-expert users to create a detailed and personalized virtual copy of themselves, which can be animated with realistic clothing deformations. 
PINA does not require complete scans, nor does it require a prior learned from large datasets of clothed humans. 
Learning a complete avatar in this setting is challenging, since only few depth observations are available, which are noisy and incomplete (i.e. only partial visibility of the body per frame).
We propose a method to learn the shape and non-rigid deformations via a pose-conditioned implicit surface and a deformation field, defined in canonical space.
This allows us to fuse all partial observations into a single consistent canonical representation.  
Fusion is formulated as a global optimization problem over the pose, shape and skinning parameters. 
The method can learn neural avatars from real noisy RGB-D sequences for a diverse set of people and clothing styles and these avatars can be animated given unseen motion sequences.

\setlength{\topsep}{0pt}
\setlength{\parskip}{.5ex}
\renewcommand{\floatsep}{1ex}
\renewcommand{\textfloatsep}{1ex}
\renewcommand{\dblfloatsep}{1ex}
\renewcommand{\dbltextfloatsep}{1ex}

\end{abstract}

\section{Introduction}
\vspace{-0.1cm}
Making immersive AR/VR a reality requires methods to effortlessly create personalized avatars. %
Consider telepresence as an example: the remote participant requires means to simply create a detailed scan of themselves and the system must then be able to re-target the avatar in a realistic fashion to a new environment and to new poses.
Such applications impose several challenging constraints: 
\begin{inparaenum}[i)]
 \item to generalize to unseen users and clothing, no specific prior knowledge such as template meshes should be required
 \item the acquired 3D surface must be animatable with realistic surface deformations driven by complex body poses
 \item the capture setup must be unobtrusive, ideally consisting of a single consumer-grade sensor  (e.g. Kinect)
 \item the process must be automatic and may not require technical expertise, rendering traditional skinning and animation pipelines unsuitable.
\end{inparaenum}  
To address these requirements, we introduce a novel  method for learning Personalized Implicit Neural Avatars (\methodname) from only a sequence of monocular RGB-D video.

Existing methods do not fully meet these criteria. 
Most state-of-the-art dynamic human models~\cite{burov2021dsfn,loper2015smpl,CAPE:CVPR:20,bhatnagar2020ipnet} represent humans as a parametric mesh and deform it via linear blend skinning (LBS) and pose correctives. Sometimes learned displacement maps to capture details of tight-fitting clothing are used~\cite{burov2021dsfn}. 
However, the fixed topology and resolution of meshes limit the type of clothing and dynamics that can be captured. 
To address this, several methods~\cite{saito2019pifu, chibane20ifnet} propose to learn neural implicit functions to model static clothed humans. Furthermore, several methods that learn a neural avatar for a specific outfit from watertight meshes ~\cite{Saito:CVPR:2021,chen2021snarf,tiwari2021neural,deng2020nasa,Peng_2021_ICCV,liu2021neural,habermann2021}  have been proposed. These methods either require complete full-body scans with accurate surface normals and registered poses~\cite{Saito:CVPR:2021,chen2021snarf,tiwari2021neural,deng2020nasa} or rely on complex and intrusive multi-view setups~\cite{Peng_2021_ICCV,liu2021neural,habermann2021}.

Learning an animatable avatar from a monocular RGB-D sequence is challenging since raw depth images are noisy and only contain partial views of the body (\figref{fig:teaser}, left). 
At the core of our method lies the idea to fuse partial depth maps into a single, consistent representation and to learn the articulation-driven deformations at the same time. 
To do so, we formulate an implicit signed distance field (SDF) in canonical space. 
To learn from posed observations, the inverse mapping from deformed to canonical space is required. We follow SNARF~\cite{chen2021snarf} and locate the canonical correspondences via optimization.
A key challenge brought on by the monocular RGB-D setting is to learn from incomplete point clouds.
 Inspired by \emph{rigid} learned SDFs for objects \cite{gropp2020implicit}, we propose a point-based supervision scheme that enables learning of articulated \emph{non-rigid} shapes (\ie clothed humans). Transforming the spatial gradient of the SDF into posed space and comparing it to surface normals from depth images leads to the learning of geometric details.
Training is formulated as a global optimization that jointly optimizes the canonical SDF, the skinning fields and the per-frame pose. \methodname learns animatable avatars without requiring any additional supervision or priors extracted from large datasets of clothed humans.

In detailed ablations, we shed light on the key components of our method. We compare to existing methods in the reconstruction and animation tasks,
showing that our method performs best across several datasets and settings. 
Finally, we demonstrate the ability to capture and animate different humans in a variety of clothing styles qualitatively. 

In summary, our contributions are:
\begin{compactitem}
 \item a method to fuse partial RGB-D observations into a canonical, implicit representation of 3D humans; and
 \item to learn an animatable SDF representation directly from partial point clouds and normals; and
 \item a formulation which jointly optimizes shape, per-frame pose and skinning weights.
\end{compactitem}

\noindent The code and video will be available on the project page:
\href{https://zj-dong.github.io/pina/}{\color{magenta}{https://zj-dong.github.io/pina/}}.

\section{Related Work}
\figurePipeline

\paragraph{Parametric Models for Clothed Humans}
A large body of literature utilizes explicit surface representations (particularly polygonal meshes) for human body modeling~\cite{anguelov2005scape, hasler2009statistical, joo2018total,pavlakos2019expressive,xu2020ghum,osman2020star, bhatnagar2019mgn}. These works typically leverage parametric models for minimally clothed human bodies~\cite{pavlakos2019expressive,dong2021shape,song2020lgd,kocabas2020vibe,bogo2016keep,fang2021reconstructing} (e.g. SMPL~\cite{loper2015smpl}) and use a displacement layer on top of the minimally clothed body to model clothing~\cite{guo2021human, alldieck2019learning,alldieck2018video,ma2020learning, neophytou2014layered,yang2018analyzing}. Recently, DSFN~\cite{burov2021dsfn} proposes to embed MLPs into the canonical space of SMPL to model pose-dependent deformations. 
However, such methods depend upon SMPL learned skinning for deformation and are upper-bounded by the expressiveness of the template mesh.
During animation or reposing, the surface deformations of parametric human models rely on the skinning weights trained from minimally clothed body scans~\cite{loper2015smpl, ma2020learning,Zhang_2017_CVPR}. These methods suffer from artifacts during reposing as they rely on skinning weights of a naked body for animation which may be incorrect for points on the surface of the garment. In contrast, our method represents clothed humans as a flexible implicit neural surface and jointly learns shape and a neural skinning field from depth observations. Other methods \cite{gundogdu2019garnet,guan2012drape, patel20tailornet, santesteban2021self,Bertiche_2021_ICCV} leverage physical simulation to drape garments onto the SMPL model. These approaches are a promising direction towards more realistic cloths deformation compared to previous template-based methods with fixed skinning weights. These methods are orthogonal to ours: we focus on acquiring a surface representation of body and clothing from the raw inputs, without assuming prior knowledge about the subject.

\vspace{-0.4cm}

\paragraph{Implicit Human Models from 3D Scans}
Implicit neural representations~\cite{park2019deepsdf,mescheder2019occupancy,chen2019learning} can handle topological changes better~\cite{bozic2021neural, park2021hypernerf} and have been used to reconstruct clothed human shapes~\cite{huang2020arch,li2020monocular,saito2019pifu,saito2020pifuhd,he2020geo,peng2021neural,raj2021anr,chen2022gdna, xiu2022icon}. Typically, based on a learned prior from large-scale datasets, they recover the geometry of clothed humans from images~\cite{saito2019pifu,saito2020pifuhd,zheng2021pamir,xiu2022icon} or point clouds~\cite{chibane20ifnet}. However, these reconstructions are static and cannot be reposed. Follow-up work \cite{huang2020arch, bhatnagar2020ipnet} attempts to endow static reconstructions with human motion based on a generic deformation field which tends to output unrealistic animated results. To model pose-dependent clothing deformations, SCANimate~\cite{Saito:CVPR:2021} proposes to transform scans to canonical space in a weakly supervised manner and to learn the implicit shape model conditioned on joint-angle rotations. 
Follow-up works further improve the generalization ability to unseen poses and accelerate the training process via a displacement network~\cite{tiwari2021neural}, deform the shape via a forward warping field~\cite{chen2021snarf, zheng2021avatar} or leverage prior information from large-scale human datasets~\cite{wang2021metaavatar}. However, all of these methods require complete and registered 3D human scans for training, even if they sometimes can be fine-tuned on RGB-D data. In contrast, \methodname is able to learn a personalized implicit neural avatar directly from a short monocular RGB-D sequence without requiring large-scale datasets of clothed human 3D scans or other priors.

\vspace{-0.4cm}
\paragraph{Reconstructing Clothed Humans from RGB-D Data} 
One straightforward approach to acquiring a 3D human model from RGB-D data is via per-frame reconstruction~\cite{chibane20ifnet,bhatnagar2020ipnet}. To achieve this, IF-Net \cite{chibane20ifnet} learns a prior to reconstruct an implicit function of a human and IP-Net \cite{bhatnagar2020ipnet} extends this idea to fit SMPL to this implicit surface for articulation. However, since input depth observations are partial and noisy, artifacts appear in unseen regions. Real-time performance capture methods incrementally fuse observations into a volumetric SDF grid. DynamicFusion~\cite{newcombe2015dynamicfusion} extends earlier approaches for static scene reconstruction~\cite{newcombe2011kinectfusion} to non-rigid objects. BodyFusion \cite{BodyFusion} and DoubleFusion \cite{DoubleFusion} build upon this concept by incorporating an articulated motion prior and a parametric body shape prior. Follow-up work~\cite{bozic2021neural,li2021posefusion,burov2021dsfn} leverages a neural network to model the deformation or to refine the shape reconstruction. However, it is important to note that such methods only reconstruct the surface, and sometimes the pose, for tracking purposes but typically do not allow for the acquisition of skinning information which is crucial for animation.  
In contrast, our focus differs in that we aim to acquire a detailed avatar including its surface and skinning field for reposing and animation.

\section{Method}
We introduce \methodname, a method for learning personalized neural avatars from a single RGB-D video, illustrated in~\figref{fig:pipeline}.
At the core of our method lies the idea to fuse partial depth maps into a single, consistent representation of the 3D human shape and to learn the articulation-driven deformations at the same time via global optimization.

We parametrize the 3D surface of clothed humans as a pose-conditioned implicit signed-distance field (SDF) and a learned deformation field in canonical space (Sec.~\ref{sec:avatar_model}). This parametrization enables the fusion of partial and noisy depth observations. This is achieved by transforming the canonical surface points and the spatial gradient into posed space, enabling supervision via the input point cloud and its normals. 
Training is formulated as global optimization (Sec.~\ref{sec:training}) to jointly optimize the per-frame pose, shape and skinning fields without requiring prior knowledge extracted from large datasets. Finally, the learned skinning field can be used to articulate the avatar (Sec.~\ref{sec:animation}).

\subsection{Implicit Neural Avatar}
\label{sec:avatar_model}

\paragraph{Canonical Representation}

We model the human avatar in canonical space and use a neural network $f_{\text{sdf}}$ to predict the signed distance value for any 3D point $\mathbf{x}_c$ in this space. To model pose-dependent local non-rigid deformations such as wrinkles on clothes, we concatenate the human pose $\mathbf{p}$ as additional input and model $f_{\text{sdf}}$ as:
\begin{equation}
    f_{\text{sdf}}: \mathbb{R}^{3} \times \mathbb{R}^{n_{p}} \rightarrow \mathbb{R} .
\end{equation}
The pose parameters ($\mathbf{p}$) are defined consistently to the SMPL skeleton~\cite{loper2015smpl} and $n_{p}$ is their dimensionality. 
The canonical shape $\mathcal{S}$ is then given by the zero-level set of $f_{\text{sdf}}$:
\begin{equation}
     \mathcal{S} = \{ \mathbf{\ x}_c \ |\ f_{\text{sdf}}(\mathbf{x}_c,\mathbf{p}) = 0 \ \}
\end{equation}
In addition to signed distances, we also compute normals in canonical space. We empirically find that this resolves high-frequency details better than calculating the normals in the posed space. The normal for a point $\mathbf{x}_c$ in canonical space is computed as the spatial gradient of the signed distance function at that point (attained via backpropagation):
\begin{equation}
    \mathbf{n}_c = \nabla_{\mathbf{x}} f_{\text{sdf}}(\mathbf{x}_c, \mathbf{p}) .
\end{equation}
To deform the canonical shape into novel body poses we additionally model deformation fields. 
To animate implicit human shapes in the desired body pose $\mathbf{p}$, we leverage linear blend skinning (LBS).
The skeletal deformation of each point in canonical space is modeled as the weighted average of a set of bone transformations $\mathbf{B}$,
which are derived from the body pose $\mathbf{p}$.
We follow \cite{chen2021snarf} and define the skinning field in  canonical space using a neural network $f_{w}$ to model the continuous LBS weight field:

\setlength{\abovedisplayskip}{7pt}
\setlength{\belowdisplayskip}{7pt}

\begin{equation}
f_{w}: \mathbb{R}^{3}  \rightarrow  \mathbb{R}^{n_b} .
\end{equation}
Here, $n_b$ denotes the number of joints in the transformation and $\mathbf{w}_c= \{w_c^1,...,w_c^{n_b}\} = f_w(\mathbf{x}_c)$  represents the learned skinning weights for $\mathbf{x}_c$.

\paragraph{Skeletal Deformation}

Given the bone transformation matrix $\mathbf{B}_i$ for joint $i \in \{1,...,n_b\}$, a canonical point $\mathbf{x_c}$ is mapped to the deformed point $\mathbf{x}_d$ as follows:
\begin{equation}
\label{eq:5}
    \mathbf{x}_d = \sum_{i = 1}^{n_b} w_{c}^i \mathbf{B}_i \, \mathbf{x}_c 
\end{equation}
The normal of the deformed point $\mathbf{x_d}$ in posed space is calculated analogously:
\begin{equation}
    \mathbf{n}_d = \sum_{i = 1}^{n_b}w_{c}^i\, \mathbf{R}_i \, \mathbf{n}_c
\end{equation}
where $\mathbf{R}_i$ is the rotation component of $\mathbf{B}_i$.

To compute the signed distance field $SDF(\mathbf{x}_d)$ in deformed space, we need the canonical correspondences $\mathbf{x}_c^*$.

\paragraph{Correspondence Search}\label{sec:search}
For a deformed point $\mathbf{x}_d$, we follow \cite{chen2021snarf} and compute its canonical correspondence set $\mathbf{\mathcal{X}}_c = \{\mathbf{x}_c^{1},...,\mathbf{x}_c^{k}\}$, which contains $k$ canonical candidates satisfying Eq.~\ref{eq:5}, via an iterative root finding algorithm. Here, $k$ is an empirically defined hyper-parameter of the root finding algorithm (see \suppmat for more details).

Note that due to topological changes, there exist one-to-many mappings when retrieving canonical points from a deformed point, \ie, the same point $\mathbf{x}_d$ may correspond to multiple different valid $\mathbf{x}_c$. Following  Ricci et al.~\cite{ricci1973constructive} , we composite these proposals of implicitly defined surfaces into a single SDF via the union (minimum) operation:
\begin{equation}
SDF(\mathbf{x}_d) = \min_{\mathbf{x}_c\in \mathbf{\mathcal{X}}_c} f_{\text{sdf}}(\mathbf{x}_c)    
\end{equation}
The canonical correspondence $\mathbf{x}_c^{*}$ is then given by:
\begin{equation}
\mathbf{x}_c^* = \argmin_{\mathbf{x}_c\in \mathbf{\mathcal{X}}_c} f_{\text{sdf}}(\mathbf{x}_c)   
\end{equation}

\subsection{Training Process}
\label{sec:training}

Defining our personalized implicit model in canonical space is crucial to integrating partial observations across all depth frames because it provides a common reference frame. Here, we formally describe this fusion process. We train our model jointly \wrt body poses and the weights of the 3D shape and skinning networks.

\paragraph{Objective Function}
\label{sec:3.2}
Given an RGB-D sequence with N input frames, we minimize the following objective:
\begin{equation}
\begin{aligned}
\mathcal{L}(\mathbf{\Theta}) = & \sum_{i=1}^{N}\mathcal{L}_{\text{on}}^{i}(\mathbf{\Theta})
+\lambda_{\text{off}} \mathcal{L}_{\text{off}}^{i}(\mathbf{\Theta})+\lambda_{\text{eik}} \mathcal{L}_{\text{eik}}^{i}(\mathbf{\Theta}) \\
\end{aligned}
\end{equation}
$\mathcal{L}_{\text{on}}^{i}$ represents an on-surface loss defined on the human surfaces for frame $i$.  $\mathcal{L}_{\text{off}}^{i}$ represents the off-surface
loss which helps to carve free-space and $\mathcal{L}_{\text{eik}}^{i}$ is the Eikonal regularizer which ensures a valid signed distance field. $\mathbf{\Theta}$ is the set of optimized parameters which includes the shape network weights $\mathbf{\Theta_{\text{sdf}}}$, the skinning network weights $\mathbf{\Theta}_{w}$ and the pose parameters $\mathbf{p}_i$ for each frame. 

To calculate $\mathcal{L}_{\text{on}}^{i}$, we first back-project the depth image into 3D space to obtain partial point clouds $\mathbf{\mathcal{P}}_{\text{on}}^{i}$ of human surfaces for each frame $i$. For each point $\mathbf{x}_d$ in $\mathbf{\mathcal{P}}_{\text{on}}^{i}$, we additionally calculate its corresponding normal $\mathbf{n}_d^{\text{obs}}$ from the raw point cloud using principal component analysis of points in a local neighborhood. $\mathcal{L}_{\text{on}}^{i}$ is then defined as

\begin{equation}
\begin{aligned}
\label{eq:9}
  \mathcal{L}_{\text{on}}^{i} &= \lambda_{\text{sdf}} \,\mathcal{L}_{\text{sdf}}^{i} + \lambda_{n}\, \mathcal{L}_{\text{n}}^{i}\\
  &= \lambda_{\text{sdf}}\sum_{\mathbf{x}_d\in \mathbf{\mathcal{P}}_{\text{on}}^{i}}|SDF(\mathbf{x}_d)|  + \lambda_{n}\sum_{\mathbf{x}_d\in \mathbf{\mathcal{P}}_{\text{on}}^{i}} \|NC(\mathbf{x}_d)\|
\end{aligned}
\end{equation}
Here, $NC(\mathbf{x}_d) = \mathbf{n}_d^{\text{obs}}(\mathbf{x}_d)-\mathbf{n}_d(\mathbf{x}_d)$.

We add two additional terms to regularize the optimization process. $\mathcal{L}_{\text{off}}^{i}$ complements $\mathcal{L}_{\text{on}}^{i}$ by randomly sampling points $\mathbf{\mathcal{P}}_{\text{off}}^{i}$ that are far away from the body surface. For any point $\mathbf{x}_d$ in $\mathbf{\mathcal{P}}_{\text{off}}^{i}$, we calculate the signed distance between this point and an estimated body mesh (see initialization section below). This signed distance $SDF_{body}(\mathbf{x}_d)$ serves as pseudo ground truth to force plausible off-surface SDF values. $\mathcal{L}_{\text{off}}^{i}$ is then defined as:
\begin{equation}
\label{eq:10}
\mathcal{L}_{\text{off}}^{i} = \sum_{\mathbf{x}_d \in \mathbf{\mathcal{P}}_{\text{off}}^{i}} |SDF(\mathbf{x}_d) - SDF_{body}(\mathbf{x}_d)|
\end{equation}
Following IGR \cite{gropp2020implicit}, we leverage $\mathcal{L}_{\text{eik}}^{i} $ to force the shape network $f_{\text{sdf}}$ to satisfy the Eikonal equation in canonical space:
\begin{equation}
    \mathcal{L}_{\text{eik}}^{i} =  \mathbb{E}_{\mathbf{x}_c}\left(\|\nabla f_{\text{sdf}}(\mathbf{x}_c)\| -1\right)^{2}
\end{equation}

\paragraph{Implementation}
The implicit shape network and blend skinning network are implemented as MLPs. We use positional encoding \cite{mildenhall2020nerf} for the query point $\mathbf{x}_c$ to increase the expressive power of the network. We leverage the implicit differentiation derived in \cite{chen2021snarf} to compute gradients during iterative root finding. 

\vspace{-0.05cm}

\paragraph{Initialization}
We initialize body poses by fitting SMPL model~\cite{loper2015smpl} to RGB-D observations. This is achieved by  minimizing the distances from point clouds to the SMPL mesh and jointly minimizing distances between the SMPL mesh and the corresponding surface points obtained from a  DensePose~\cite{guler2018densepose} model.
Please see \suppmat. for details.

\vspace{-0.05cm}

\paragraph{Optimization}
 Given a sequence of RGB-D video, we deform our neural implicit human model for each frame based on the 3D pose estimate and compare it with its corresponding RGB-D observation.
 This allows us to jointly optimize both shape parameters  $\mathbf{\Theta}_{\text{sdf}}$, $\mathbf{\Theta}_{w}$ and pose parameters $\mathbf{p}_i$ of each frame and makes our model robust to noisy initial pose estimates. 
 We follow a two-stage optimization protocol for faster convergence and more stable training: First, we pretrain the shape and skinning networks in canonical space based on the SMPL meshes obtained from the initialization process. Then we optimize the shape network, skinning network and poses jointly to match the RGB-D observations.

\subsection{Animation}
\label{sec:animation}

To generate animations, we discretize the deformed space at a pre-defined resolution and estimate $SDF(\mathbf{x}_d)$ for every point $\mathbf{x}_d$ in this grid via correspondence search (Sec \ref{sec:search}). We then extract meshes via  MISE~\cite{mescheder2019occupancy}. %

\newcommand{\tablebuff}{
\begin{table}[t]
\centering
\begin{tabular}{lccc}

\hline  Method & $\mathbf{IoU} \uparrow$ & $\mathbf{C}-\ell_{2} \downarrow$ & $\mathbf{N C} \uparrow$ \\
\hline
IP-Net \cite{bhatnagar2020ipnet}  & $0.783$ & $2.1 \mathrm{}$ & $0.861$ \\
CAPE \cite{CAPE:CVPR:20}  & $0.648$ & $2.5 \mathrm{}$ & $0.844$ \\
DSFN \cite{burov2021dsfn} & ${0 . 8 3 2}$ & ${1.6} \mathrm{}$ & $0 . 9 1 6$ \\

\hline
  Ours & $\mathbf{0 . 8 7 9}$ & $\mathbf{1.1} \mathbf{}$ & $\mathbf{0 . 9 2 7}$ \\
\hline

\end{tabular}
\caption{\textbf{Quantitative evaluation on BUFF.} We provide rendered depth maps as input for all methods. Our method consistently outperforms all other baselines in all metrics (see \figref{fig:buff_result} for qualitative comparison).}
\label{tab:buff}
\end{table}
}

\newcommand{\tableablationopt}{
\begin{table}[t]
\centering
\begin{tabular}{lccc}

\hline   Method & $\mathbf{IoU} \uparrow$ & $\mathbf{C}-\ell_{2} \downarrow$ & $\mathbf{N C} \uparrow$ \\

\hline
  Ours w/o pose opt. & $0.850$ & $1.6 \mathrm{}$ & $0.887$ \\
\hline
  Ours & $\mathbf{0 . 8 7 9}$ & $\mathbf{1.1} \mathbf{}$ & $\mathbf{0 . 9 2 7}$ \\
\hline

\end{tabular}
\caption{\textbf{Importance of pose optimization on BUFF.} We evaluate the reconstruction results of our method without jointly optimizing pose and shape. }
\label{tab:ablation_opt}
\end{table}
}

\newcommand{\tablecape}{
\begin{table}[t]
\begin{tabular}{lcccc}
\hline Method & Input &$\mathbf{I o U} \uparrow$ & $\mathbf{C}-\ell_{2} \downarrow$ & $\mathbf{N C} \uparrow$ \\
\hline 
IP-Net \cite{bhatnagar2020ipnet} &3D & $0.916$ & $0.786 $ & $0.843$ \\
SCANimate \cite{Saito:CVPR:2021} &3D & $\mathbf{0.941}$ & $\mathbf{0.596} $ & $\mathbf{0.906}$ \\

\hline 
SCANimate \cite{Saito:CVPR:2021} &2.5D & ${0.665}$ & ${3.704} \mathrm{}$ & $0.785 $\\

Ours & 2.5D &$\mathbf{0.946}$ & $\mathbf{0.666} \mathbf{}$ & $\mathbf{0.906 }$ \\
\hline
\end{tabular}
\caption{\textbf{Quantitative evaluation on CAPE.} Our method outperforms IP-Net and SCANimate (2.5D) by a large margin and achieves comparable result with SCANimate (3D) which is trained on complete 3D meshes, a significantly easier setting compared to using partial 2.5D data as input. }

\vspace{-1em}
\label{tab:cape}
\end{table}
}

\newcommand{\tableablationcomponent}{
\begin{table}
\begin{tabular}{lccc}
\hline Method  & $\mathbf{I o U} \uparrow$ & $\mathbf{C}-\ell_{2} \downarrow$ & $\mathbf{N C} \uparrow$ \\
\hline Ours (w/o pose cond.) &$0.936$ & $0.991 \mathrm{}$ & $0.884$  \\
Ours (w/ SMPL weights) & $0.945$  & $0.643$ & $0.887$ \\
\hline
Ours (complete) & $\mathbf{0.955}$ & $\mathbf{0.604}$ & $\mathbf{0.912}$ \\
\hline
\end{tabular}
\caption{\textbf{Importance of pose-dependent deformations and learned skinning weights on CAPE.} We evaluate the animation results of our method without pose conditioning and driven by SMPL skinning weights.}
\label{tab:ablation_avatar}
\end{table}
}

\section{Experiments}

We first conduct ablations on our design choices. Next, we compare our method with state-of-the-art approaches on the reconstruction and animation tasks. Finally, we demonstrate  personalized avatars learned from only a single monocular RGB-D video sequence qualitatively.

\subsection{Datasets}
We first conduct experiments on two standard datasets with clean scans projected to RGB-D images to evaluate our performance on both \textbf{reconstruction} and \textbf{animation}. To further demonstrate the robustness of our method to real-world sensor noise, we collect a dataset with a single Kinect including various challenging garment styles.

\vspace{-0.2cm}

\paragraph{BUFF Dataset \cite{Zhang_2017_CVPR}:} This dataset contains textured 3D scan sequences. Following \cite{burov2021dsfn}, we obtain monocular RGB-D data by rendering the scans and use them for our \textbf{reconstruction} task, comparing to the ground truth scans. 

\vspace{-0.2cm}

\paragraph{CAPE Dataset \cite{CAPE:CVPR:20}:} This dataset contains registered 3D meshes of people wearing different clothes while performing various actions. It also provides corresponding ground-truth SMPL parameters. Following \cite{Saito:CVPR:2021}, we conduct \textbf{animation} experiments on CAPE. To adapt CAPE to our monocular depth setting, we acquire single-view depth inputs by rendering the meshes. The most challenging subject (blazer) is used for evaluation where 10 sequences are used for training and 3 unseen sequences are used for evaluating the \textbf{animation} performance. Note that our method requires RGB-D for initial pose estimation since CAPE does not provide texture, we take the ground-truth poses for training (same for the baselines).

\vspace{-0.2cm}

\paragraph{Real Data:} To show robustness and generalization of our method to noisy real-world data, we collect RGB-D sequences with an Azure Kinect at $30$ fps (each sequence is approximately 2-3 minutes long). We use the RGB images for pose initialization. We learn avatars from this data and animate avatars with unseen poses \cite{aist-dance-db, AMASS:ICCV:2019,CAPE:CVPR:20}. 

\vspace{-0.2cm}

\paragraph{Metrics:}We consider volumetric IoU, Chamfer distance (cm) and normal consistency for evaluation.

\figureablationopt
\subsection{Ablation Study}

\paragraph{Joint Optimization of Pose and Shape:}

\label{sec:ablation_opt}
The initial pose estimate from a monocular RGB-D video is usually noisy and can be inaccurate. To evaluate the importance of \emph{jointly} optimizing pose and shape, we compare our full model to a version without pose optimization. \textbf{Results}: \tabref{tab:ablation_opt} shows that joint optimization of pose and shape is crucial to achieve high reconstruction quality and globally accurate alignment (Chamfer distance and IoU). It is also important to recover details (normal consistency). As shown in \figref{fig:ablation_opt}, unnatural reconstructions such as the artifacts on the head and trouser leg can be corrected by pose optimization. %

\tableablationopt

\paragraph{Deformation Model:}

The deformation of the avatar can be split into \textbf{pose-dependent deformation} and skeletal deformation via LBS with the learned skinning field. To model pose-dependent deformations such as cloth wrinkles, we leverage a pose-conditioned shape network to represent the SDF in canonical space. \textbf{Results}:  \figref{fig:ablation_deform} shows that without pose features, the network cannot represent dynamically changing surface details of the blazer, and defaults to a smooth average. This is further substantiated by a 70$\%$ increase in Chamfer distance, compared to our full method.
\vspace{-0.2cm}

To show the importance of \textbf{learned skinning weights}, we compare our full model to a variant with a fixed  shape network (w/ SMPL weights). Points are deformed using SMPL blend weights at the nearest SMPL vertices. \textbf{Results}: \tabref{tab:ablation_avatar} indicates that our method outperforms the baseline in all metrics. In particular, the normal consistency improved significantly.
This is also illustrated in \figref{fig:ablation_deform}, where the baseline (w/ SMPL weights) is noisy and yields artifacts. This can be explained by the fact that the skinning weights of SMPL are defined only at the mesh vertices of the naked body, thus they can't model complex deformations.

\figureablationdeform
\figureBUFF
\tableablationcomponent

\subsection{Reconstruction Comparisons}\label{sec:ipnet}
\vspace{-0.2cm}
\tablebuff 
\label{sec:recon}

\paragraph{Baselines:}
Although not our primary goal, we also compare to several reconstruction methods, including IP-Net \cite{bhatnagar2020ipnet}, CAPE \cite{CAPE:CVPR:20} and DSFN \cite{burov2021dsfn}. The experiments are conducted on the BUFF \cite{Zhang_2017_CVPR} dataset. The RGB-D inputs are rendered from a sequence of registered 3D meshes. IP-Net relies on a learned prior from \cite{twindom, treedys}. It takes the partial 3D point cloud from each depth frame as input and predicts the implicit surface of the human body. The SMPL+D model is then registered to the reconstructed surface. DSFN models per-vertex offsets to the minimally-clothed SMPL body via pose-dependent MLPs. For CAPE, we follow the protocol in DSFN~\cite{burov2021dsfn} and optimize the latent codes based on the RGB-D observations. %

\paragraph{Results:}
Tab.~\ref{tab:buff} summarizes the reconstruction comparison on BUFF. 
We observe that our method leads to better reconstructions in all three metrics compared to current SOTA methods. A qualitative comparison is shown in \figref{fig:buff_result}. Compared to methods based on implicit reconstruction, i.e., IP-Net, our method reconstructs person-specific details better and generates complete human bodies. This is due to the fact that IP-Net reconstructs the human body frame-by-frame and can't leverage information across the sequence. In contrast, our method solves the problem via global optimization. Compared to methods with explicit representations, i.e., CAPE and DSFN, our method reconstructs details (hair, trouser leg) that geometrically differ from the minimally clothed human body better. We attribute this to the flexibility of implicit shape representations. %

\figureCape

\subsection{Animation Comparisons}
\vspace{-0.1cm}
\paragraph{Baselines:}
We compare animation quality on CAPE~\cite{CAPE:CVPR:20} with IP-Net~\cite{bhatnagar2020ipnet} and SCANimate~\cite{Saito:CVPR:2021} as baselines. IP-Net does not natively fuse information across the entire depth sequence (discussed in Sec \ref{sec:ipnet}). For a fair comparison, we feed one complete T-pose scan of the subject as input to IP-Net and predict implicit geometry and leverage the registered SMPL+D model to deform it to unseen poses. For SCANimate, we create two baselines. The first baseline (SCANimate 3D) is learned from complete meshes and follows the original setting of SCANimate. Note that in this comparison ours is at a disadvantage since we only assume monocular depth input without accurate surface normal information. Therefore, we also compare to a variant (SCANimate 2.5D) which operates on equivalent 2.5D inputs.

\figureQualRes

\paragraph{Results:}
Tab.~\ref{tab:cape} shows the quantitative results. Our method outperforms IP-Net and SCANimate (2.5D), and achieves comparable results to SCANimate (3D) which is trained on complete and noise-free 3D meshes.  \figref{fig:cape_result} shows that the clothing deformation of the blazer is unrealistic when animating  IP-Net. This may be due to overfitting to the training data. Moreover, the animation is driven by skinning weights that are learned from minimally-clothed human bodies. As seen in \figref{fig:cape_result}, SCANimate also leads to unrealistic animation results for unseen poses. This is because the deformation field in SCANimate depends on the pose of the \emph{deformed} object, which limits generalization to unseen poses \cite{chen2021snarf}. Furthermore, we find that this issue is amplified in SCANimate (2.5D) with partial point clouds. In contrast, our method solves this problem well via jointly learned skinning field and shape in canonical space. 

\tablecape
\subsection{Real-world Performance}
\vspace{-0.1cm}
To demonstrate the performance of our method on noisy real-world data, we show results on additional RGB-D sequences from an Azure Kinect in \figref{fig:qualitative_result}. More specifically, we learn a neural avatar from an RGB-D video and drive the animation using unseen motion sequences from ~\cite{aist-dance-db,AMASS:ICCV:2019,CAPE:CVPR:20}. Our method is able to reconstruct complex cloth geometries like hoodies, high collar and puffer jackets. Moreover, we demonstrate reposing to novel out-of-distribution motion sequences including dancing and exercising.

\section{Conclusion}
In this paper, we presented \methodname to learn personalized implicit avatars for reconstruction and animation from noisy and partial depth maps. The key idea is to represent the implicit shape and the pose-dependent deformations in canonical space which allows for fusion over all frames of the input sequence. We propose a global optimization that enables joint learning of the skinning field and surface normals in canonical representation.
Our method learns to recover surface details and is able to animate the human avatar in novel unseen poses. We compared the method to explicit and neural implicit state-of-the-art baselines and show that we outperform all baselines in all metrics. Currently, our method does not model the appearance of the avatar. This is an exciting direction for future work. We discuss potential negative societal impact and limitations in the \suppmat.

{\small
\noindent\textbf{Acknowledgements:} Zijian Dong was supported by ELLIS. Xu Chen was supported by the Max Planck ETH Center for Learning Systems. 
}

{\small
\bibliographystyle{ieee_fullname}
\bibliography{egbib}
}

\end{document}